\documentclass[runningheads]{llncs}

\usepackage{graphicx}
\usepackage{url}
\usepackage{rotating}
\usepackage{multirow}
\usepackage[ruled,vlined,linesnumbered,algo2e]{algorithm2e}

\begin{document}

\title{Towards Exploiting Implicit Human Feedback for Improving RDF2vec Embeddings\thanks{Copyright © 2020 for this paper by its authors. Use permitted under Creative Commons License Attribution 4.0 International (CC BY 4.0).}}
\titlerunning{Exploiting Implicit Human Feedback for Improving RDF2vec Embeddings}

\author{Ahmad Al Taweel \and Heiko Paulheim}

\institute{University of Mannheim, Germany\\Data and Web Science Group\\\email{ahmad.altaweel@outlook.de,heiko@informatik.uni-mannheim.de}}

\maketitle              

\begin{abstract}
RDF2vec is a technique for creating vector space embeddings from an RDF knowledge graph, i.e., representing each entity in the graph as a vector.
It first creates sequences of nodes by performing random walks on the graph.
In a second step, those sequences are processed by the word2vec algorithm for creating the actual embeddings.
In this paper, we explore the use of external edge weights for guiding the random walks.
As edge weights, transition probabilities between pages in Wikipedia are used as a proxy for the human feedback for the importance of an edge.
We show that in some scenarios, RDF2vec utilizing those transition probabilities can outperform both RDF2vec based on random walks as well as the usage of graph internal edge weights.

\end{abstract}

\section{Introduction}
RDFvec \cite{ristoski2016rdf2vec} was originally conceived for exploiting Semantic Web knowledge graphs in data mining. Since most popular data mining tools require a feature vector representation of records, various techniques have been proposed for creating vector space representations from subgraphs, including straightforward techniques like adding features for datatype properties or binary dimensions for types, \cite{ristoski2014comparison}, as well as techniques based on graph kernels \cite{losch2012graph,de2013fast}.

Given the increasing popularity of the word2vec family of word embedding techniques \cite{mikolov2013efficient}, which learns feature vectors for words based on the context in which they appear, this approach has been proposed to be transferred to graphs as well. Since word2vec operates on (word) sequences, several approaches have been proposed which first turn a graph into sequences by performing random walks, and then applying the idea of word2vec to those sequences. Such approaches include node2vec \cite{grover2016node2vec}, DeepWalk \cite{perozzi2014deepwalk}, Wembedder \cite{nielsen2017wembedder}, and the aforementioned RDF2vec.

While random walks are a straightforward technique for transforming a graph into sequences, they lack a notion of importance of edges, and assign each edge in the graph -- be it important or not -- the same weight. This observation has lead to the inspection of various biases in the walk generation, e.g., preferring edges to nodes with a high or low PageRank \cite{cochez2017biased}. However, the results using those graph-internal biases were not very conclusive so far.

In this paper, we present an initial set of experiments which, instead of generating a bias signal from data which is \emph{internal} to the graph (e.g., by utilizing PageRank), uses \emph{external} data for assigning weights to edges. More specifically, we utilize transition probabilities between Wikipedia pages, created from user logs in Wikipedia, as a proxy for the importance of an edge in a knowledge graph derived from Wikipedia, i.e., DBpedia.

The rest of this paper is structured as follows. In section 2, we discuss some relevant related work. In section 3, we introduce the approach and data used. We present some initial experimental findings in section 4. In section 5, we discuss the results and their possible impact on applications, as well as how they can be generalized to other datasets and embedding methods.

\section{Related Work}
The generation of knowledge graph embeddings is a highly active and vibrant field, with many new approaches being proposed at a very high pace~\cite{wang2017knowledge}. Despite the high number of approaches that have been proposed, little light has been shed on assigning a weight or importance measure to edges when constructing the embedding space.

One such approach is the extension of RDF2vec already discussed \cite{cochez2017biased} above, which replaces \emph{random} by \emph{biased} walks, where the bias comes from property frequencies or PageRank of target nodes. 

In \cite{mai2018support}, an information theoretic approach is taken to assign weights to edges. The authors argue that the importance of an edge in the graph can be determined by the likelihood that it can be reconstructed by inference. The more easily it gets reconstructed, the more redundant it is. Consequently, the authors propose an extension of translational embeddings which focus more on the less redundant edges.

Both approaches use \emph{graph-internal} sources to generate the weights, i.e., the weights are derived directly from the graph without any additional use of external data. In contrast, we propose an approach for using external data for deriving such weights, thereby adding a truly fresh signal to the embedding approach.

\section{Approach}
Our approach is a direct extension of RDF2vec. As discussed above, RDF2vec uses random walks to generate sequences of nodes and edges, and then applies word2vec on top of those sequences for generating feature vectors. Both variants of word2vec -- CBOW and SkipGram\footnote{CBOW (Context Bag of Words) tries to predict a token in a sequence given its surrounding tokens, while SkipGram tries the opposite, i.e., predicting the surroundings of a token in a sequence given that token.} -- have been implemented for RDF2vec, with the latter often showing superior results \cite{ristoski2019rdf2vec}.

In \cite{cochez2017biased}, edge weights have been introduced as an extension to RDF2vec. The adaptation of the RDF2vec is shown in Algorithm~\ref{algo:randomWalk}. When generating weighted random walks, the next edge to follow is selected in the function \emph{selectEdge} (line 11) using a probability distribution computed from the edge weights of all connecting edges of a node, i.e., the probability of each edge $e_{ij}$ from node \emph{i} to node \emph{j} being followed is the weight of that edge, divided by the sum of the weights of all outgoing edges of \emph{i}:
\begin{equation}
    Pr\left [ e_{ij} \right ]=\frac{weight(e_{ij})}{\sum_{e_{ik}}weight(e_{ik}) }
\end{equation}
Note that the classic RDF2vec implementation of RDF2vec is equivalent to using uniform weights, i.e., all weights being set to 1.

\begin{algorithm2e}[t]
	\SetAlgoLined
	\KwData{$G = (V, E)$: RDF Graph, $d$: walk depth, $n$: number of walks}
	\KwResult{$P_G$: Set of sequences}
		$P_G$ := $\emptyset$\\
	\ForEach{vertex $v \in V$}{
		$n_v$ = n\\
		\While{$n_v$ $>$ 0}{
				$w$ = initialize walk\\
				add $v$ to $w$\\
				$currentVertex$ = $v$\\
				$d_v$ = $d$\\
				\While{$d_v$ $>$ 0}{
					$E_c$ = $currentVertex.outEdges()$\\
					$e$ = selectEdge($E_c$)\\
					$d_v$ = $d_v$ - 1\\
					add $e$ to $w$\\
						\If{$d_v$ $>$ 0}
						{
							$v_e$ = $e.endVertex()$\\
							add $v_e$ to $w$\\
							$currentVertex$ = $v_e$\\
							$d_v$ = $d_v$ - 1\\
						}
					}
					add $w$ to $P_G$\\
					$n_v$ = $n_v$ - 1\\
				}
		}
	\caption{Algorithm for generating weighted RDF graph walks \cite{ristoski2019rdf2vec}}
	\label{algo:randomWalk}
\end{algorithm2e}

In this paper, we generate RDF2vec embeddings for DBpedia \cite{auer2007dbpedia}, a knowledge graph extracted from Wikipedia. In DBpedia, each entity corresponds to a Wikipedia page. Hence, we can utilize information generated for Wikipedia also for computations on DBpedia.

For assigning weights to edges, we collect implicit human feedback. 
To that end, we utilize transition likelihoods in user navigation, generated from Wiki\-pedia's usage logs\footnote{\url{https://meta.wikimedia.org/wiki/Research:Wikipedia_clickstream}}. Those represent the number of link transitions (i.e., clicks) from a Wikipedia page to another one.
This transitional probability is used as implicit user feedback, serving as a proxy for a human rating for the importance of an edge in the knowledge graph.

\begin{figure}[t]
    \centering
    \includegraphics[width=\textwidth]{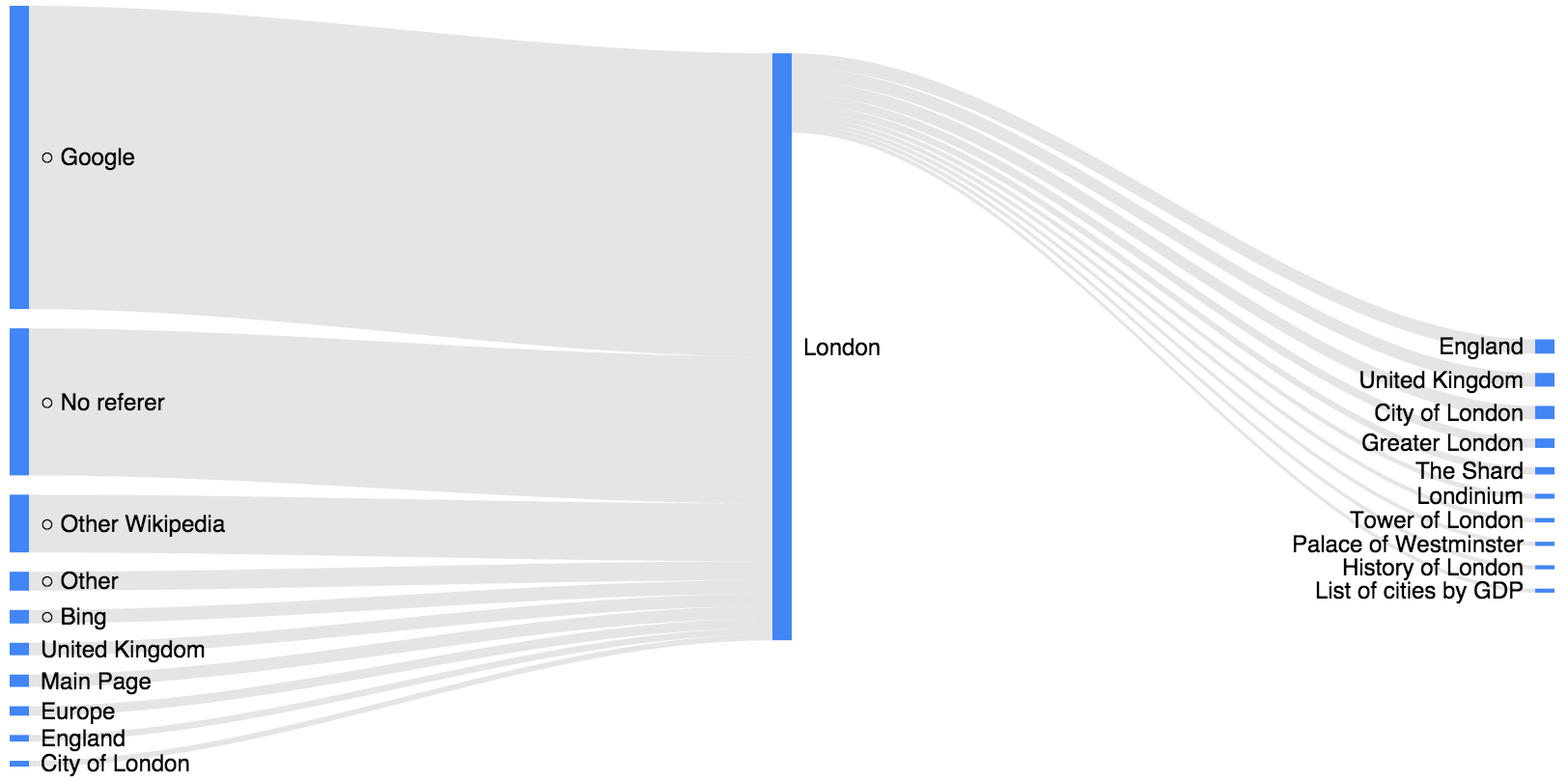}
    \caption{Clickstream data example, showing the top 10 click sources and targets for the Wikipedia page for London. Source: \protect\url{https://meta.wikimedia.org/wiki/Research:Wikipedia_clickstream}}
    \label{fig:example}
\end{figure}

Figure~\ref{fig:example} shows an example excerpt of the clickstream data for the Wikipedia page London. On the left hand side, the top pages from which a user gets to the Wikipedia page for London are depicted: the top five are Google and Bing, as well as other Web pages, the most likely Wikipedia pages are United Kingdom, main Page, Europe, England, and City of London. On the right hand side, the top 10 pages to which a user navigates from the London pages are depicted. Each transition has a probability, indicated by the strength of the connecting line.

\begin{figure}[t]
    \centering
    \includegraphics[width=\textwidth]{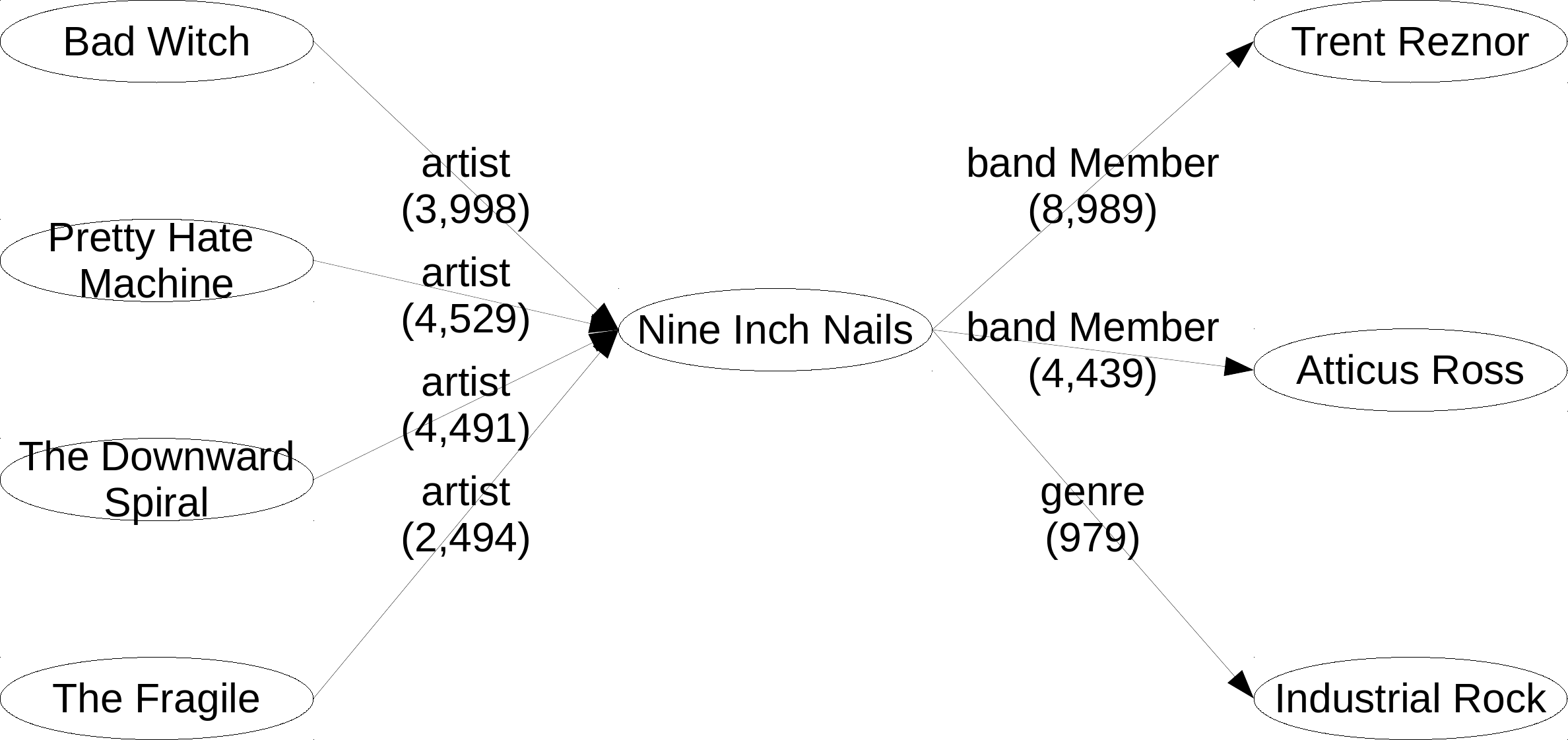}
    \caption{Excerpt from DBpedia, using transition likelihoods from the Clickstream dataset}
    \label{fig:excerpt}
\end{figure}

Note that not every link in Wikipedia corresponds to a statement in DBpedia; numbers not referring to Wikipedia pages (e.g., links from/to non-Wikipedia pages) are ignored in the formula above generating the random walks.
This means that the transition probabilities are normalized so that the sum of probabilities of a transition from one Wikipedia page to all Wikipedia pages linked from this page is $1$.
Fig.~\ref{fig:excerpt} shows an excerpt of DBpedia, with weights extracted from the Wikipedia Clickstream dataset. In this case, the sequence
\begin{verbatim}
Pretty_Hate_Machine artist Nine_Inch_Nails bandMember Trent_Reznor
\end{verbatim}
would have a much higher probability to be generated than the sequence
\begin{verbatim}
Bad_Witch artist Nine_Inch_Nails genre Industrial_Rock
\end{verbatim}

\section{Experiments}
For evaluating the impact of weights in the RDF2vec embeddings, we conduct two series of experiments. First, we follow the setup in \cite{pellegrino2019configurable}, solving a couple of entity classification and regression tasks based on a set of benchmarks for machine learning on the semantic web \cite{ristoski2016collection}. For those datasets, three sets of entities -- cities, movies, and albums -- have been augmented with an external variable to be predicted, i.e., the quality of living index for cities, and the metacritic score for movies and albums.

Second, we perform a set of experiments with a content-based recommender system based on RDF2vec embeddings \cite{ristoski2019rdf2vec}. 
Here, we use a variant of the MovieLens dataset, which collects user ratings for movies, which has been linked to DBpedia.

In both sets of experiments, we compare the results achieved using our weighted RDF2vec embeddings generated on DBpedia against the standard implementation of RDF2vec, as well as the three best performing variants using graph internal weighting schemes, as reported in \cite{cochez2017biased}, i.e.,
\begin{description}
    \item[Predicate Frequency] uses the global frequency of a predicate as an edge weight. Consequently, edges with frequent properties are followed more frequently than edges with infrequent properties.
    \item[PageRank] uses the PageRank of the target node of an edge as the edge weight. Hence, nodes which are more central in the graph are more likely to be part of walks.
    \item[Inverse PageRank] does the opposite, as it uses the inverse of the PageRank of an object as the edge weight. Here, walks are more likely to contain nodes which are less central in the graph.
\end{description}
In particular, the latter two strategies are very different: PageRank creates walks which mainly contain central nodes, i.e., it creates a very uneven distribution reflecting the popularity of nodes, whereas inverse PageRank tries to favor less central nodes and create a more even distribution in which central nodes are not treated with preference, i.e., long tail entities are covered better.

The code used for the experiments, including all the variants that were considered for comparison, is available online.\footnote{\url{https://github.com/ataweel55/RDF2VEC}}

\begin{table}[t]
    \centering
    \caption{Classification results (accuracy)}
    \label{tbl:classification}
    \scriptsize
\begin{tabular}{|c|| c| c| c| c|| c| c| c| c|| c| c| c| c|} 
 \hline
  \multirow{2}{*}{Strategy/Dataset} & \multicolumn{4}{c||}{Cities} & \multicolumn{4}{c||}{Metacritic Movies} & \multicolumn{4}{c|}{Metacritic Albums} \\
\cline{2-13}
 & NB & KNN & SVM & C4.5 & NB & KNN & SVM & C4.5& NB & KNN & SVM & C4.5 \\ [0.5ex] 
 \hline\hline
  
Uniform SG 200w 200v 4d & 73.25 & 72.90 & 76.32 & 50.26
                        & 72.29 & 75.81 & 75.94 & 67.28 
                        & 71.46 & 72.87 & 68.66 & 63.45 \\ [1ex] 
 \hline
Uniform SG 500w 200v 4d & 59.25 & 67.51 & 73.01 & 61.27  
                        & 65.25 & 79.62 & 81.19 & 74.52 
                        & 69.25 & 72.85 & 75.80 & 64.92  \\[1ex] 
 \hline
Uniform SG 500w 200v 8d & 71.65 & 75.52 & 72.82 & 59.48 
                        & 76.41 & 72.35 & \textbf{83.51} & 76.83 
                        & 76.36 & 75.96 & \textbf{81.72} & 67.25  \\[1ex] 
 \hline
Uniform SG 500w 500v 8d & 87.63 & 70.15 & 82.70 & 68.72
                        & 81.35 & 76.24 & 83.35 & 71.52
                        & 75.64 & 75.92 & 78.63 & 67.18  \\[1ex] 
 \hline
 \hline
Predicate frequency weight SG 200w 200v 4d & 72.15 & 70.77 & 73.37 & 51.25 
                                            & 74.60 & 75.21 & 76.13 & 72.42 
                                            & 69.82 & 71.90 & 73.53 & 62.22 \\ [1ex] 
                                            
 \hline
Page-Rank weight SG 200w 200v 4d & 74.16 & 72.73 & 67.21 & 62.83 
                                 & 71.32 & 69.19 & 79.70 & 74.43 
                                 & 69.61 & 71.98 & 74.47 & 68.11 \\[1ex] 
 \hline
Inverse Page-Rank Weight SG 200w 200v 4d & 73.98 & 69.37 & 74.60 & 57.38
                                         & 68.68 & 71.25 & 72.71 & 64.93 
                                         & 67.39 & 73.57 & 64.27 & 58.66 \\ [1ex] 
 \hline
 \hline
Click-Stream weight CBOW 200w 200v 4d & 60.25 & 64.90 & 74.32 & 65.26
                                      & 67.29 & 75.81 & 77.94 & 69.20 
                                      & 71.15 & 72.38 & 74.47 & 63.48 \\  [1ex] 
 \hline
Click-Stream weight SG 200w 200v 4d & 74.68 & 71.20 & 77.94 & 53.45
                                    & 71.11 & 77.31 & 79.52 & 67.78 
                                    & 70.46 & 72.20 & 67.18 & 62.88 \\ [1ex] 
 \hline
Click-Stream weight SG 500w 200v 4d & 61.22 & 73.90 & 78.39 & 62.45 
                                     & 65.14 & 80.49 & 81.44 & 79.65 
                                     & 71.19 & 69.78 & 74.15 & 69.37 \\ [1ex] 
 \hline
Click-Stream weight SG 500w 500v 8d & \textbf{88.63} & 69.70 & 81.62 & 71.93
                                & 82.44 & 74.69 & 82.27 & 69.12
                                & 77.63 & 72.84 & 79.48 & 66.81 \\ [1ex] 
 \hline
\end{tabular}

\end{table}

\begin{table}[t]
    \centering
    \caption{Regression results (Root Mean Squared Error)}
    \label{tbl:regression}
    \scriptsize
\begin{tabular}{|c|| c| c| c|| c| c| c|| c| c| c| } 
 \hline
  \multirow{2}{*}{Strategy/Dataset} & \multicolumn{3}{c||}{Cities} & \multicolumn{3}{c||}{Metacritic Movies} & \multicolumn{3}{c|}{Metacritic Albums} \\
\cline{2-10}
 & LR & KNN & M5 & LR & KNN & M5 & LR & KNN & M5 \\ [0.5ex] 
 \hline\hline
Uniform SG 200w 200v 4d & 16.64 & 15.92 & 16.87  
                        & 17.16 & 19.33 & 17.79  
                        & 12.38 & 14.42 & 12.97  \\ [1ex] 
 \hline
Uniform SG 500w 200v 4d & 14.74 & 12.78 & 14.52
                        & 16.95 & 18.62 & 17.28  
                        & 13.56 & 14.33 & 13.65   \\[1ex] 
                        
 \hline
Uniform SG 500w 200v 8d & 13.68 & 14.93 & 13.45 
                        & 16.79 & 18.40 & 16.97
                        & 13.17 & 14.36 & 13.22  \\[1ex] 
                        
 \hline
Uniform SG 500w 500v 8d & 12.23 & 13.81 & 10.86 
                        & 16.42 & 18.03 & 16.68 
                        & 12.48 & 13.63 & 11.96  \\[1ex] 
                        
 \hline
 \hline
 Predicate frequency weight SG 200w 200v 4d & 16.54 & 17.83 & 17.56
                                    & 18.28 & 20.90 & 19.72 
                                    & 14.31 & 16.88 & 13.44 \\ [1ex] 
                                        
 \hline
Page-Rank weight SG 200w 200v 4d & 14.74 & 14.57 & 16.14 
                                 & 17.63 & 20.81 & 16.86 
                                 & 12.57 & 15.72 & 12.56 \\[1ex] 
 \hline
Inverse Page-Rank weight SG 200w 200v 4d & 14.87 & 16.59 & 14.93
                                       & 16.10 & 18.44 & 16.16
                                      & 11.56 & 12.93 & 11.51 \\[1ex] 
    
 \hline
 \hline
Click-Stream weight CBOW 200w 200v 4d & 15.13 & 13.64 & 16.13 
                                       & 19.76 & 20.91 & 19.48 
                                       & 13.52 & 14.29 & 13.58 \\ [1ex] 
 \hline
Click-Stream weight SG 200w 200v 4d & 15.48 & 16.16 & 15.24 
                                     & 17.31 & 19.27 & 16.78 
                                     & 12.28 & 13.75 & 12.21 \\ [1ex] 
 \hline
Click-Stream weight SG 500w 200v 4d & 13.96 & 16.53 & 14.82 
                                     & \textbf{15.66} & 16.15 & 15.86 
                                     & 11.70 & 12.50 & 11.77 \\ [1ex]
 \hline
Click-Stream weight SG 500w 500v 8d & 12.25 & 12.57 & \textbf{10.11} 
                                     & 15.72 & 16.89 & 15.80 
                                     & 10.79 & 11.67 & \textbf{11.14} \\ [1ex] 
 \hline
\end{tabular}
\end{table}

\subsection{Results on Machine Learning Benchmark Datasets}
In the three benchmark datasets used, the goal is to predict the quality of living in cities, and the metacritic score of movies and albums. For classification, those are discretized (into high and low), for regression, the goal is to predict the actual number. In those experiments, we follow the setup in the original RDF2vec paper \cite{ristoski2016rdf2vec}, utilizing four standard out of the box classification algorithms (Naive Bayes, nearest neighbors, Support Vector Machines, and C4.5 decision trees) and three regression algorithms (linear regression, nearest neighbors, and M5 regression trees).\footnote{We are aware that there might be better performing state of the art algorithms, e.g., deep neural networks or XGboost. However, for our goal of comparing different RDF2vec variants, we tried to follow the original evaluation protocol of RDF2vec as closely as possible.} 

Tables~\ref{tbl:classification} and~\ref{tbl:regression} depict the results for the classification and regression tasks. It can be observed that for the classification tasks, the embeddings incorporating clickstream data outperform the default RDF2vec embeddings in one out of three tasks, while for the regression, they outperform those in all three cases. Moreover, we can observe that the embeddings utilizing the external signal for the weighting schemes are superior to all three techniques utilizing graph internal signals.

\subsection{Results on Recommendation Datasets}
In this experiment, we use the RDF2vec graph embedding method in the framework of content-based recommender systems for feature generation and rely on a comparatively easy suggestion algorithm, i.e., the items are recommended using the K-Nearest Neighbors technique with cosine similarity in the RDF2vec embedding space. Formally, this technique evaluates the proximity of objects by means of cosine similarity between the respective RDF2vec vectors and then selects a subset of those – the neighbors – for each object, which will be used to predict the user u a rating for a fresh item I as follows: 

\begin{equation}
r^*\left ( u,i \right ) = \frac{\sum_{j \in ratedItems\left ( u \right )} cosineSim(j,i).r_{u,j})}{\sum_{j \in ratedItems\left ( u \right )}|cosineSim(j,i)|}
\end{equation}

Where $ratedItems(u)$ is a collection of products already assessed by the user $u, r_{u, j}$ suggests the user $u$ use $j$ rating and $cosineSim(j, i)$ shows the cosine similarity score between $j$ and $i$ products. The size of the considered neighborhood is restricted to 5 in our experiments. 

\begin{table}[t]
\centering
\caption{Results of the ItemKNN approach on Movielens dataset }
\label{tbl:results_recommendation}
\scriptsize
\begin{tabular}{ |c|| r| r| r| } 

\hline
Strategy & Precision & Recall & F1\\
\hline
\hline
Uniform SG 200w 200v 4d & 0.05128 & 0.02466 & 0.03330 \\
\hline
Uniform SG 500w 200v 4d & 0.04852 & 0.03024 & 0.03725 \\
\hline
Uniform SG 500w 200v 8d & 0.04279 & 0.02612 & 0.03243 \\
\hline
Uniform SG 500w 500v 8d & 0.02692 & 0.01624 & 0.02025 \\
\hline
\hline
Predicate frequency weight SG 200w 200v 4d & 0.01946 & 0.0960 & 0.03236 \\
\hline
Page-Rank weight SG 200w 200v 4d & 0.03251 & 0.01828 & 0.02340 \\
\hline
Inverse Page-Rank weight SG 200w 200v 4d & 0.03924 & 0.02369 & 0.02954 \\
\hline
 \hline
Click-Stream weight CBOW 200w 200v 4d & 0.03162 & 0.01348 & 0.01890 \\
\hline
Click-Stream weight SG 200w 200v 4d & \textbf{0.05261} & \textbf{0.03625} & \textbf{0.04292} \\
\hline
Click-Stream weight SG 500w 200v 4d & 0.04622 & 0.02573 & 0.03305 \\
 \hline
Click-Stream weight SG 500w 500v 8d & 0.02489 & 0.01925 & 0.02170 \\
\hline
\end{tabular}
\end{table}

Table~\ref{tbl:results_recommendation} depicts the results of the recommender system experiments. It can be observed that the results incorporating human feedback outperform both the default RDF2vec approach as well as the three variants using graph-internal weights.

\section{Conclusion and Outlook}
In this paper, we have introduced an approach for incorporating an external signal -- in our case: page transition probabilities in Wikipedia -- for creating knowledge graph embeddings. We have shown that the performance of RDF2vec models can be increased by exploiting such a signal for edge weights.

So far, we have carried out experiments on DBpedia, and used edge weights derived from Wikipedia. The same technique would be applicable for knowledge graphs based on Wikipedia, such as YAGO~\cite{suchanek2008yago} and CaLiGraph~\cite{heist2019uncovering}, as well as knowledge graphs based on other Wikis~\cite{hertling2018dbkwik} for which log files are available. For other knowlede graphs and datasets, other methods of obtaining edge weights need to be investigated. Moreover, for Wikipedia-based knowledge graphs, other measures of weights than interaction histories might be feasible (e.g., assessing the importance of a link based on the text of a Wikipedia page, or analyzing the edit history of a page to find out how early the link to the other page was added).

RDF2vec is not the only knowledge graph embedding technique in which edge weights can be exploited. In~\cite{mai2018support}, the authors argue that translational embedding techniques such as TransE~\cite{bordes2013translating} and its descendants can also be adapted in a way that they use edge weights. So far, this has only done by using graph internal weights. Hence, given our observations that external weights work better for RDF2vec than graph internal weights, we want to investigate the usage of graph external weights in such embedding techniques as well.

Since RDF2vec has been used in quite a few downstream applications, we want to investigate the effect of graph external edge weights in such application as well. Besides recommender systems, possible fields are the use of RDF2vec for ontology alignment \cite{portisch2018alod2vec}, for analyzing changes in ontologies \cite{jurisch2018rdf2vec}, or reconciling information extracted from different texts \cite{alam2017reconciling}. Moreover, given a specific task, it would be interesting to investigate to which extent task-specific weighting schemes might be utilized.

\bibliographystyle{splncs04}
\bibliography{references.bib}

\end{document}